# Wrap-Up: a Trainable Discourse
# Module for Information Extraction

**Stephen Soderland**                                    SODERLAN@CS.UMASS.EDU
**Wendy Lehnert**                                        LEHNERT@CS.UMASS.EDU
*Department of Computer Science, University of Massachusetts*
*Amherst, MA 01003-4610*

## Abstract

The vast amounts of on-line text now available have led to renewed interest in information extraction (IE) systems that analyze unrestricted text, producing a structured representation of selected information from the text. This paper presents a novel approach that uses machine learning to acquire knowledge for some of the higher level IE processing. Wrap-Up is a trainable IE discourse component that makes intersentential inferences and identifies logical relations among information extracted from the text. Previous corpus-based approaches were limited to lower level processing such as part-of-speech tagging, lexical disambiguation, and dictionary construction. Wrap-Up is fully trainable, and not only automatically decides what classifiers are needed, but even derives the feature set for each classifier automatically. Performance equals that of a partially trainable discourse module requiring manual customization for each domain.

## 1. Introduction

An information extraction (IE) system analyzes unrestricted, real world text such as newswire stories. In contrast to information retrieval systems which return a pointer to the entire document, an IE system returns a structured representation of just the information from within the text that is relevant to a user's needs, ignoring irrelevant information.

The first stage of an IE system, sentence analysis, identifies references to relevant objects and typically creates a case frame to represent each object. The second stage, discourse analysis, merges together multiple references to the same object, identifies logical relationships between objects, and infers information not explicitly identified by sentence analysis. The IE system operates in terms of domain specifications that predefine what types of information and relationships are considered relevant to the application. Considerable domain knowledge is used by an IE system: about domain objects, relationships between objects, and how texts typically describe these objects and relationships.

Much of the domain knowledge can be automatically acquired by corpus-based techniques. Previous work has centered on knowledge acquisition for some of the lower level processing such as part-of-speech tagging and lexical disambiguation. N-gram statistics have been highly successful in part-of-speech tagging (Church, 1988; DeRose, 1988). Weischedel (1993) has used corpus-based probabilities both for part-of-speech tagging and to guide parsing. Collocation data has been used for lexical disambiguation by Hindle (1989), Brent (1993), and others. Examples from a training corpus have driven both part-of-speech and semantic tagging (Cardie, 1993) and dictionary construction (Riloff, 1993).

This paper describes Wrap-Up (Soderland & Lehnert, 1994), the first system to automatically acquire domain knowledge for the higher level processing associated with discourse analysis. Wrap-Up uses supervised learning to induce a set of classifiers from a training corpus of representative texts, where each text is accompanied by hand-coded target output. We implemented Wrap-Up with the ID3 decision tree algorithm (Quinlan, 1986), although other machine learning algorithms could have been selected.

Wrap-Up is a fully trainable system and is unique in that it not only decides what classifiers are needed for the domain, but automatically derives the feature set for each classifier. The user supplies a definition of the objects and relationships of interest to the domain and a training corpus with hand-coded target output. Wrap-Up does the rest with no further hand coding needed to tailor the system to a new domain.

Section 2 discusses the IE task in more detail, introduces the microelectronics domain, and gives an overview of the CIRCUS sentence analyzer. Section 3 describes Wrap-Up, giving details of how ID3 trees are constructed for each discourse decision, how features are automatically derived for each tree, and requirements for applying Wrap-Up to a new domain. Section 4 shows the performance of Wrap-Up in two domains and compares its performance to that of a partially trainable discourse component. In Section 5 we draw some conclusions about the contribution of this research. A detailed example from the microelectronics domain is given in an appendix.

## 2. The Information Extraction Task

This section gives an overview of information extraction and illustrates IE processing with a sample text fragment from the microelectronics domain. We then discuss the need for trainable IE components to acquire knowledge for a new domain.

### 2.1 An Overview of IE

An information extraction system operates at two levels. First, sentence analysis identifies information that is relevant to the IE application. Then discourse analysis, which we will focus on in this paper, takes the output from sentence analysis and assembles it into a coherent representation of the entire text. All of this is done according to predefined guidelines that specify what objects from the text are relevant and what relationships between objects are to be reported.

Sentence analysis can be further broken down into several stages, each applying different types of domain knowledge. The lowest level is preprocessing, which segments the text into words and sentences. Each word is assigned a part-of-speech tag and possibly a semantic tag in preparation for further processing. Different IE systems will do varying amounts of syntactic parsing at this point. Most research sites that participated in the ARPA-sponsored Message Understanding Conferences (MUC-3, 1991; MUC-4, 1992; MUC-5, 1993) found that robust, shallow analysis and pattern matching performed better than more elaborate, but brittle, parsing techniques.

The CIRCUS sentence analyzer (Lehnert, 1990; Lehnert et al., 1992) does shallow syntactic analysis to identify simple syntactic constituents, and to distinguish active and passive voice verbs. This shallow syntactic analysis is sufficient for the extraction task, which uses local linguistic patterns to instantiate case frames, called concept nodes (CN's) used by CIRCUS.

Each CN definition has a trigger word and a syntactic pattern relative to that word. Whenever the trigger word occurs in the text, CIRCUS looks in one of the syntactic buffers for appropriate information to extract. Some CN definitions will extract information from the subject or from the direct object, first testing for active or passive voice. Other CN definitions look for a prepositional phrase with a particular preposition. Examples of CN extraction patterns from a particular domain are shown in Section 2.3.

Discourse analysis starts with the output from the sentence analyzer, in this case a set of concept nodes representing locally extracted information. Other work on discourse has often involved tracking shifts in topic and in the speaker/writer's goals (Grosz & Sidner, 1986; Liddy et al., 1993) or in resolving anaphoric references (Hobbs, 1978). Discourse processing in an IE system may concern itself with some of these issues, but only as a means to its main objective of transforming bits and pieces of extracted information into a coherent representation.

One of the first tasks of discourse analysis is to merge together multiple references to the same object. In a domain where company names are important, this will involve recognizing the equivalence of a full company name ("International Business Machines, Inc.") with shortened forms of that name ("IBM") and generic references ("the company", "the U.S. computer maker"). Some manually engineered rules seem unavoidable for coreference merging. Another example is merging a domain object with a less specific reference to that object. In the microelectronics domain a reference to "DRAM" chips may be merged with a reference to "memory" or an "I-line" process merged with "lithography."

Much of the work of discourse analysis is to identify logical relationships between extracted objects, represented as pointers between objects in the output. Discourse analysis must also be able to infer missing objects that are not explicitly stated in the text and in some cases split an object into multiple copies or discard an object that was erroneously extracted.

The current implementation of Wrap-Up begins discourse processing after coreference merging has been done by a separate module. This is primarily because manual engineering seems unavoidable in coreference. Work is underway to extend Wrap-Up to include all of IE discourse processing by incorporating a limited amount of domain-specific code to handle such things as company name aliases and generic references to domain objects.

Wrap-Up divides its processing into six stages, which will be described more fully in Section 3. They are:

1. Filtering out spuriously extracted information
2. Merging objects with their attributes
3. Linking logically related objects
4. Deciding when to split objects into multiple copies
5. Inferring missing objects
6. Adding default slot values

At this point an example from a specific domain might help. The following sections introduce the microelectronics domain, then illustrate sentence analysis and discourse analysis with a short example from this domain.

## 2.2 The Microelectronics Domain

The microelectronics domain was one of the two domains targetted by the Fifth Message Understanding Conference (MUC-5, 1993). According to the domain and task guidelines developed for the MUC-5 microelectronics corpus, the information to be extracted are microchip fabrication processes along with the companies, equipment, and devices associated with these processes. There are seven types of domain objects to be identified: entities (i.e. companies), equipment, devices, and four chip fabrication processes (layering, lithography, etching, and packaging).

Identifying relationships between objects is of equal importance in this domain to identifying the objects themselves. A company must be identified as playing at least one of four possible roles with respect to the microchip fabrication process: developer, manufacturer, distributor, or purchaser/user. Microfabrication processes are reported only if they are associated with a specific company in at least one of these roles. Each equipment object must be linked to a process which uses that equipment, and each device object linked to a process which fabricates that device. Equipment objects may point to a company as manufacturer and to other equipment as modules.

The following sample from the MUC-5 microelectronics domain has two companies in the first sentence, which are associated with two lithography processes from the second sentence. GCA and Sematech are developers of both the UV and I-line lithography processes, with GCA playing the additional role of manufacturer. Each lithography process is linked to the stepper equipment mentioned in sentence one.

```
GCA unveiled its new XLS stepper, which was developed with
assistance from Sematech.  The system will be available in
deep-ultraviolet and I-line configurations.
```

Figure 1 shows the five domain objects extracted by sentence analysis and the final representation of the text after discourse analysis has identified relationships between objects. Some of these relationships are directly indicated by pointers between objects. The roles that companies play with respect to a microchip fabrication process are indicated by creating a "microelectronics-capability" object with pointers to both the process and the companies.

## 2.3 Extraction Patterns

How does sentence analysis identify GCA and Sematech as company names, and extract the other domain objects such as stepper equipment, UV lithography and I-line lithography? The CN dictionary for this domain includes an extraction pattern "X unveiled" to identify company names. The subject of the active verb "unveiled" in this domain is nearly always a company developing or distributing a new device or process. However, this pattern will occasionally pick up a company that fails the domain's reportability criteria. A company that unveils a new type of chip should be discarded if the text does not specify the fabrication process.

Extracting the company name "Sematech" is more difficult since the pattern "assistance from X" is not a reliable predictor of relevant company names. There is always a trade-off between accuracy and complete coverage in deciding what extraction patterns are reliable

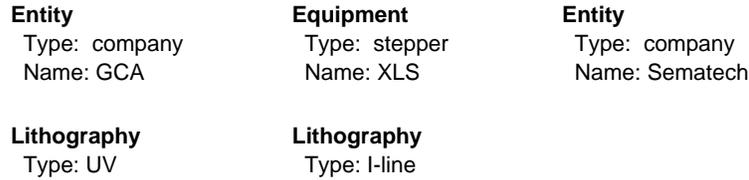

A. Five concept nodes extracted by sentence analysis.

**Entity**
Type: company
Name: GCA

**Equipment**
Type: stepper
Name: XLS

**Entity**
Type: company
Name: Sematech

**Lithography**
Type: UV

**Lithography**
Type: I-line

B. Final representation of the text after discourse analysis.

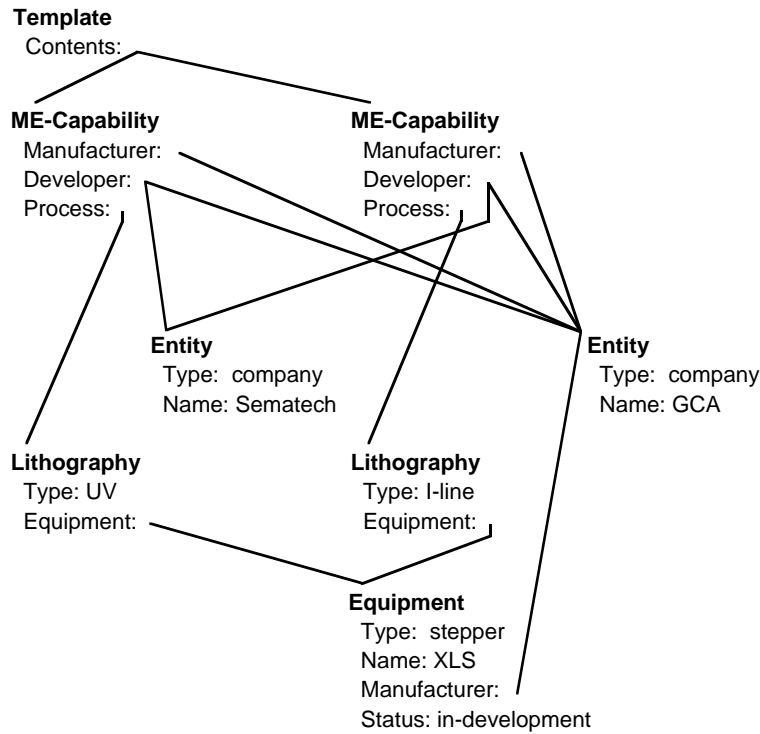

Figure 1: Output of (A) sentence analysis and (B) discourse analysis

enough to include in the CN dictionary. Including less reliable patterns increases coverage but does so at the expense of spurious extraction. The more specific pattern "developed with assistance from X" is reliable, but was missed by the dictionary construction tool (Riloff, 1993).

For many of the domain objects, such as equipment, devices, and microchip fabrication processes, the set of possible objects is predefined and a list of keywords that refer to these objects can be created. The extraction pattern "unveiled X" looks in the direct object of the active verb "unveiled", instantiating an equipment object if a keyword indicating an equipment type is found. In this example an equipment object with type "stepper" is created with the equipment name "XLS". The same stepper equipment is also extracted by the pattern "X was developed", which looks for equipment in the subject of the passive verb "developed". This equipment object is extracted a third time by the keyword "stepper" itself, which is sufficient to instantiate a stepper equipment object whether or not it occurs in a reliable extraction pattern.

The keyword "deep-ultraviolet" and the extraction pattern "available in X" are used to extract a lithography object with type "UV" from the second sentence. Another lithography object of type "I-line" is similarly extracted. Case frames are created for each of the objects identified by sentence analysis. This set of objects becomes input for the next stage of processing, discourse analysis.

## 2.4 Discourse Processing

In the full text from which this fragment comes, there are likely to be other references to "GCA" or to "GCA Corp." One of the first jobs of discourse analysis is to merge these multiple references. It is a much harder task to merge pronominal references and generic references such as "the company" with the appropriate company name. This is all part of the coreference problem that is handled by processes separate from Wrap-Up.

The main job of discourse analysis is to determine the relationships between the objects passed to it by sentence analysis. Considerable domain knowledge is needed to make these discourse-level decisions. Some of this knowledge concerns writing style, and specific phrases writers typically use to imply relationships between referents in a given domain. Is the phrase "<company> unveiled <equipment>" sufficient evidence to infer that the company is the developer of a microelectronics process? The word "unveiled" alone is not enough, since a company that unveiled a new DRAM chip may not be the developer of any new process. It may simply be using someone else's microelectronics process to produce its chip. Such inferences, particularly those about what role a company plays in a process, are often so subtle that two human analysts may disagree on the output for a given text. A human performance study for this task found that experienced analysts agreed with each other on only 80% on their text interpretations in this domain (Will, 1993).

World knowledge is also needed about the relationships possible between domain objects. A lithography process may be linked to stepper equipment, but steppers are never used in layering, etching, or packaging processes. There are delicate dependencies about what types of process are likely to fabricate what types of devices. Knowledge about the kinds of relationships typically reported in this domain can also help guide discourse processing. Stories about lithography, for example, often give the developer, manufacturer, or distributor of the process, but these roles are hardly ever mentioned for packaging processes. Companies associated with packaging tend to be limited to the purchaser/user of the packaging technology.

A wide range of domain knowledge is needed for discourse processing, some of it related to world knowledge, some to writing style. The next section discusses the need for trainable components at all levels of IE processing, including discourse analysis. Wrap-Up uses machine learning techniques to avoid months of manual knowledge engineering otherwise required to develop a specific IE application.

## 2.5 The Need for Trainable IE Components

The highest performance at the ARPA-sponsored Fifth Message Understanding Conference (MUC-5, 1993) was achieved at the cost of nearly two years of intense programming effort, adding domain-specific heuristics and domain-specific linguistic patterns one by one, followed by various forms of system tuning to maximize performance. For many real world applications, two years of development time by a team of half a dozen programmers would be prohibitively expensive. To make matters worse, the knowledge used in one domain cannot be readily transferred to other IE applications.

Researchers at the University of Massachusetts have worked to facilitate IE system development through the use of corpus-driven knowledge acquisition techniques (Lehnert et al., 1993). In 1991 a purely hand-crafted UMass system had the highest performance of any site in the MUC-3 evaluation. The following year UMass ran both a hand-crafted system and an alternate system that replaced a key component with output from AutoSlog, a trainable dictionary construction tool (Riloff, 1993). The AutoSlog variant exhibited performance levels comparable to a dictionary based on 1500 hours of manual coding. Encouraged by the success of this one trainable component, an architecture for corpus-driven system development was proposed which uses machine learning techniques to address a number of natural language processing problems (Lehnert et al., 1993). In the MUC-5 evaluation, output from the CIRCUS sentence analyzer was sent to TTG (Trainable Template Generator), a discourse component developed by Hughes Research Laboratories (Dolan, et al., 1991; Lehnert et al., 1993). TTG used machine learning techniques to acquire much of the needed domain knowledge, but still required hand-coded heuristics to turn this acquired knowledge into a fully functioning discourse analyzer.

The remainder of this paper will focus on Wrap-Up, a new IE discourse module now under development which explores the possibility of fully automated knowledge acquisition for discourse analysis. As detailed in the following sections, Wrap-Up builds ID3 decision trees to guide discourse processing and requires no hand-coded customization for a new domain once a training corpus has been provided. Wrap-Up automatically decides what ID3 trees are needed for the domain and derives the feature set for each tree from the output of the sentence analyzer.

## 3. Wrap-Up, a Trainable IE Component

This section describes the Wrap-Up algorithm, how decision trees are used for discourse analysis, and how the trees and tree features are automatically generated. We conclude with a discussion of the requirements of Wrap-Up and our experience porting to a new domain.

### 3.1 Overview

Wrap-Up is a domain-independent framework for IE discourse processing which is instantiated with automatically acquired knowledge for each new IE application. During its training phase, Wrap-Up builds ID3 decision trees based on a representative set of training texts, paired against hand-coded output keys. These ID3 trees guide Wrap-Up's processing during run time.

At run time Wrap-Up receives as input all objects extracted from the text during sentence analysis. Each of these objects is represented as a case frame along with a list of references in the text, the location of each reference, and the linguistic patterns used to extract it. Multiple references to the same object throughout the text are merged together before passing it on to Wrap-Up. Wrap-Up transforms this set of objects by discarding spurious objects, merging objects that add further attributes to an object, adding pointers between objects, and inferring the presence of any missing objects or slot values.

Wrap-Up has six stages of processing, each with its own set of decision trees designed to transform objects as they are passed from one stage to the next.

Stages in the Wrap-up Algorithm:

1. Slot Filtering

    Each object slot has its own decision tree that judges whether the slot contains reliable information. Discard the slot value from an object if a tree returns "negative".

2. Slot Merging

    Create an instance for each pair of objects of the same type. Merge the two objects if a decision tree for that object type returns "positive". This stage can merge an object with separately extracted attributes for that object.

3. Link Creation

    Consider all possible pairs of objects that might possibly be linked. Add a pointer between objects if a Link Creation decision tree returns "positive".

4. Object Splitting

    Suppose object A is linked to both object B and to object C. If an Object Splitting decision tree returns "positive", split A into two copies with one pointing to B and the other to C.

5. Inferring Missing Objects

    When an object has no other object pointing to it, an instance is created for a decision tree which returns the most likely parent object. Create such a parent and link it to the "orphan" object unless the tree returns "none". Then use decision trees from the Link Creation and Object Splitting stages to tie the new parent in with other objects.

6. Inferring Missing Slot Values

    When an object slot with a closed class of possible values is empty, create an instance for a decision tree which returns a context-sensitive default value for that slot, possibly "none".

## 3.2 Decision Trees for Discourse Analysis

A key to making machine learning work for a complex task such as discourse processing is to break the problem into a number of small decisions and build a separate classifier for each. Each of the six stages of Wrap-Up described in Section 3.1 has its own set of ID3 trees, with the exact number of trees depending on the domain specifications. The Slot Filtering stage has a separate tree for each slot of each object in the domain; the Slot Merging stage has a separate tree for each object type; the Link Creation stage has a tree for each pointer defined in the output structure; and so forth for the other stages. The MUC-5 microelectronics domain (as explained in Section 2.2) required 91 decision trees: 20 for the Slot Filtering stage, 7 for Slot Merging, 31 for Link Creation, 13 for Object Splitting, 7 for Inferring Missing Objects , and 13 for Inferring Missing Slot Values.

An example from the Link Creation stage is the tree that determines pointers from lithography objects to equipment objects. Every pair of lithography and equipment objects found in a text is encoded as an instance and sent to the Lithography-Equipment-Link tree. If the classifier returns "positive", Wrap-Up adds a pointer between these two objects in the output to indicate that the equipment was used for that lithography process.

The ID3 decision tree algorithm (Quinlan, 1986) was used in these experiments, although any machine learning classifier could be plugged into the Wrap-Up architecture. A vector space approach might seem appropriate, but its performance would depend on the weights assigned to each feature (Salton et al., 1975). It is hard to see a principled way to assign weights to the heterogeneous features used in Wrap-Up's classifiers (see Section 3.3), since some features encode attributes of the domain objects and others encode linguistic context or relative position in the text.

Let's look again at the example from Section 2.2 with the "XLS stepper" and see how Wrap-Up makes the discourse decision of whether to add a pointer from UV lithography to this equipment object. Wrap-Up encodes this as an instance for the Lithography-Equipment-Link decision tree with features representing attributes of both the lithography and equipment objects, their extraction patterns, and relative position in the text.

During Wrap-Up's training phase, an instance is encoded for every pair of lithography and equipment objects in a training text. Training instances must be classified as positive or negative, so Wrap-Up consults the hand-coded target output provided with the training text and classifies the instance as positive if a pointer is found between matching lithography and equipment objects. The creation of training instances will be discussed more fully in Section 3.4. ID3 tabulates how often each feature value is associated with a positive or negative training instance and encapsulates these statistics at each node of the tree it builds.

Figure 2 shows a portion of a Lithography-Equipment-Link tree, showing the path used to classify the instance for UV lithography and XLS stepper as positive. The parenthetical numbers for each tree node show the number of positive and negative training instances represented by that node. The *a priori* probability of a pointer from lithography to equipment in the training corpus was 34%, with 282 positive and 539 negative training instances.

ID3 uses an information gain metric to select the most effective feature to partition the training instances (p.89-90, Quinlan, 1986), in this case choosing equipment type as the test at the root of this tree. This feature alone is sufficient to classify instances with equipment type such as modular equipment, radiation source, or etching system, which have only negative instances. Apparently these types of equipment are never used by lithography processes (a useful bit of domain knowledge).

The branch for equipment type "stepper" leads to a node in the tree representing 202 positive and 174 negative training instances, raising the probability of a link to 54%. ID3 recursively selects a feature to partition each partition, in this case selecting lithography type. The branch for UV lithography leads to a partition with 27 positive and 14 negative instances, in contrast to E-beam and optical lithography which have nearly all negative instances. The next test is distance, with a value of -1 in this case since the equipment reference is one sentence earlier than lithography. This branch leads to a leaf node with 4 positive and no negative instances, so the tree returns a classification of positive and Wrap-Up adds a pointer from UV lithography to the stepper.

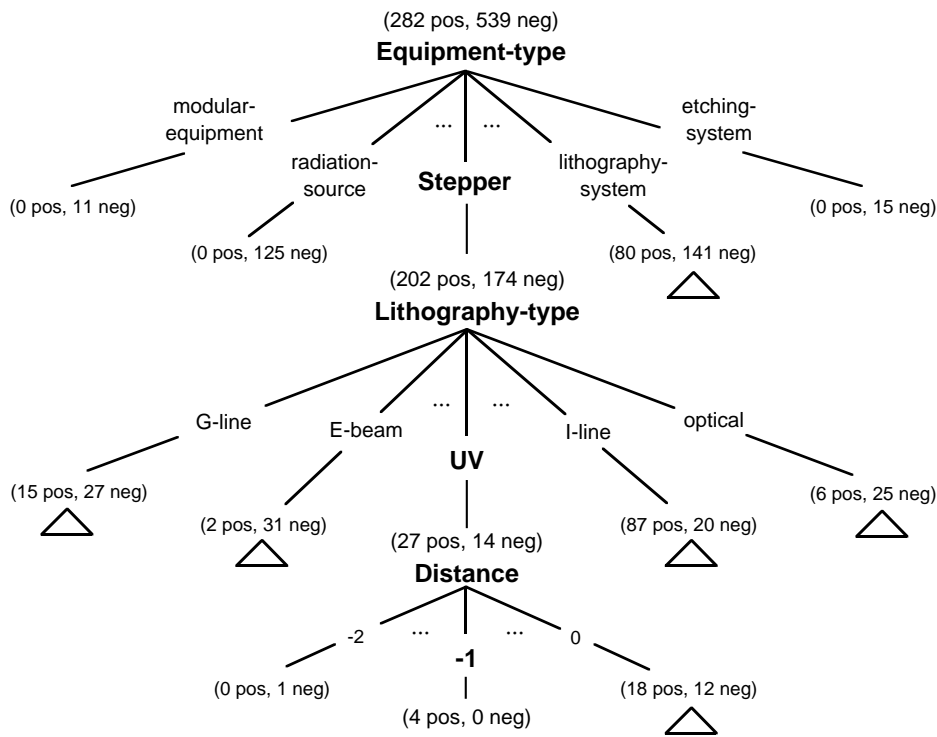

Figure 2: A decision tree for pointers from lithography to equipment objects.

This example shows how a decision tree can acquire useful domain knowledge: that lithography is never linked to equipment such as etching systems, and that steppers are often used for UV lithography but hardly ever for E-beam or optical lithography. Knowledge of this sort could be manually engineered rather than acquired from machine learning, but the hundreds of rules needed might take weeks or months of effort to create and test.

Consider another fragment of text and the tree in Figure 3 that decides whether to add a pointer from the PLCC packaging process to the ROM chip device.

```
...a new line of 256 Kbit and 1 Mbit ROM chips.  They are
available in PLCC and priced at ...
```

The instance which is to be classified by a Packaging-Device-Link tree includes features for packaging type, device type, distance between the two referents, and the extraction patterns used by sentence analysis.

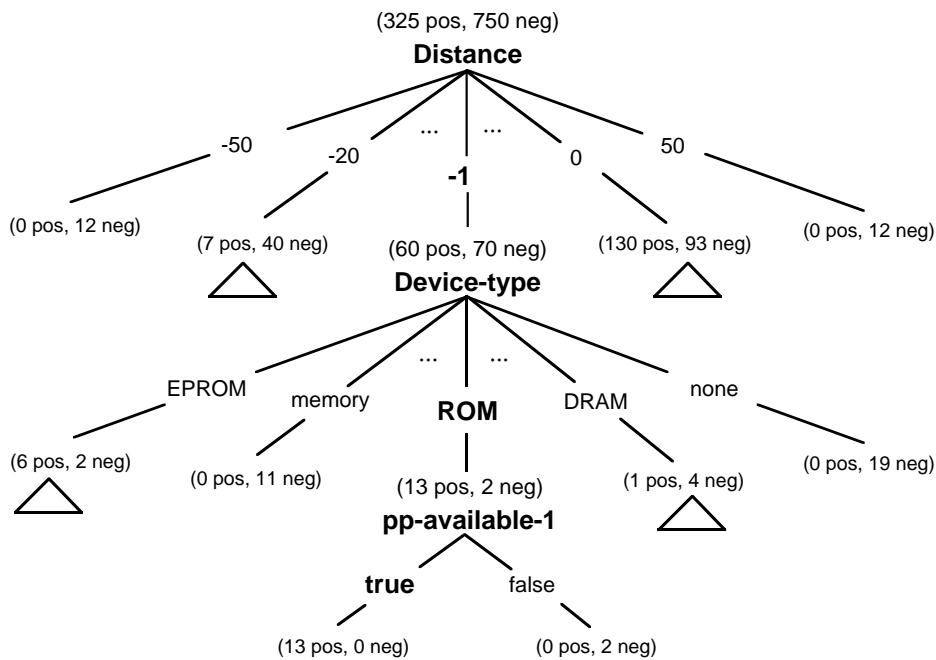

Figure 3: A tree for pointers from packaging to device objects.

ID3 selects "distance" as the root of the tree, a feature that counts the distance in sentences between the packaging and device references in the text. When the closest references were 20 or more sentences apart, hardly any of the training instances were positive. The distance is -1 in the example text, with ROM device mentioned one sentence earlier than the PLCC packaging process. As Figure 3 shows, the branch for distance of -1 is followed by a test for device type. The branch for device type ROM leads to a partition with only 15 instances, 13 positive and 2 negative. Those with PLCC packaging found in the pattern "available in X" (encoded as pp-available-1) were positive instances.

These two trees illustrate how different trees learn different types of knowledge. The most significant features in determining whether an equipment object is linked to a lithography process are real world constraints on what type of equipment can be used in lithography. This is reflected in the tree in Figure 2 by choosing equipment type as the root node followed by lithography type. There is no such overriding constraint on what type of device can be linked to a packaging technique. Here linguistic clues play a more prominent role, such as the relative position of references in the text and particular extraction patterns. The following section discusses how these linguistic-based features are encoded.

## 3.3 Generating Features for ID3 Trees

Let's look in more detail at how Wrap-Up encodes ID3 instances, using information available from sentence analysis to automatically derive the features used for each tree. Each ID3 tree handles a discourse decision about a domain object or the relationship between a pair of objects, with different stages of Wrap-Up involving different sorts of decisions.

The information to be encoded about an object comes from concept nodes extracted during sentence analysis. Concept nodes have a case frame with slots for extracted information, and also have the location and extraction patterns of each reference in the text. Consider again the example from Section 2.2.

```
GCA unveiled its new XLS stepper, which was developed with
assistance from Sematech.  The system will be available in
deep-ultraviolet and I-line configurations.
```

Sentence analysis extracts five objects from this text: the company GCA, the equipment XLS stepper, the company Sematech, UV lithography, and I-line lithography. One of several discourse decisions to be made is whether the UV lithography uses the XLS stepper mentioned in the previous sentence. Figure 4 shows the two objects that form the basis of an instance for the Lithography-Equipment-Link tree.

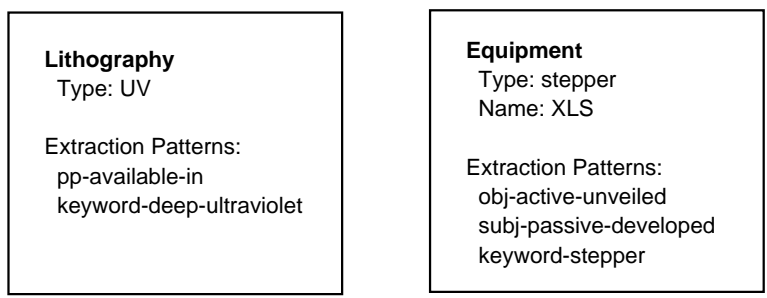

Figure 4: Two objects extracted from the sample text

Each object includes the location of each reference and the patterns used to extract them. An extraction pattern is a combination of a syntactic pattern and a specific lexical item or "trigger word" (as explained in Section 2.1). The pattern pp-available-in means that a reference to UV lithography was found in a prepositional phrase following the triggers "available" and "in".

Figure 5 shows the instance for UV lithography and XLS stepper. It encodes the attributes and extraction patterns of each object and their relative position in the text. Wrap-Up encodes each case frame slot of each object using the actual slot value for closed classes such as lithography type. Open class slots such as equipment names are encoded with the value "t" to indicate that a name was present, rather than the actual name. Using the exact name would result in an enormous branching factor for this feature and might overly influence the ID3 classification if a low frequency name happened to occur only in positive or only in negative instances.

Extraction patterns are encoded as binary features that include the trigger word and syntactic pattern in the feature name. Patterns with two trigger words such as "pp-available-in" are split into two features, "pp-available" and "pp-in". For instances that encode a pair of objects these features will be encoded as "pp-available-1" and "pp-in-1" if they refer to the first object. The count of how many such extraction patterns were used is also encoded

```
(lithography-type . UV)              (equipment-type . stepper)
(extraction-count-1 . 3)             (equipment-name . t)
(pp-available-1 . t)                 (extraction-count-2 . 3)
(pp-in-1 . t)                        (obj-unveiled-2 . t)
(keyword-deep-ultraviolet-1 . t)     (subj-passive-developed-2 . t)
                                     (keyword-stepper-2 . t)

                    (common-triggers . 0)
                    (common-phrases . 0)
                    (distance . -1)
```

Figure 5: An instance for the Lithography-Equipment-Link tree.

for each object. The feature "extraction-count" was motivated by the Slot Filtering stage since objects extracted several times are more likely to be valid than those extracted only once or twice from the text.

Another type of feature, encoded for instances involving pairs of objects, is the relative position of references to the two objects, which may be significant in determining if two objects are related. One feature easily computed is the distance in sentences between references. In this case the feature "distance" has a value of -1, since XLS stepper is found one sentence earlier than the UV lithography process. Another feature that might indicate a strong relationship between objects is the count of how many common phrases contain references to both objects. Other features list "common triggers", words included in the extraction patterns for both objects. An example of this would be the word "using" if the text had the phrase "the XLS stepper using UV technology".

It is important to realize what is *not* included in this instance. A human making this discourse decision might reason as follows. The sentence with UV lithography indicates that it is associated with "the system", which refers back to "its new XLS stepper" in the previous sentence. Part of this reasoning involves domain independent use of a definite article, and part requires domain knowledge that "system" can be a nonspecific reference to an equipment object. The current version of Wrap-Up does not look beyond information passed to it by sentence analysis and misses the reference to "the system" entirely.

Using specific linguistic patterns resulted in extremely large, sparse feature sets for most trees. The Lithography-Equipment-Link tree had 1045 features, all but 11 of them encoding extraction patterns. Since a typical instance participates in at most a dozen extraction patterns, a serious time and space bottle neck would occur if the hundreds of linguistic patterns that are *not* present were explicitly listed for each instance. We implemented a sparse vector version of ID3 that was able to efficiently handle large feature spaces by only tabulating the small number of true-valued features for each instance.

As links are added during discourse processing, objects may become complex, including many pointers to other objects. By the time Wrap-Up considers links between companies and microelectronics processes, a lithography object may have a pointer to an equipment object or to a device object, and the equipment object may in turn have pointers to other objects. Wrap-Up allows objects to inherit the linguistic context and position in the text of objects to which they point. When object A has a pointer to object B, the location and extraction patterns of references to B are treated as if they references to A. This version of inheritance is helpful, but a little too strong, ignoring the distinction between direct references and inherited references.

We have looked at the encoding of instances for isolated discourse decisions in this section. The entire discourse system is a complex series of decisions, each affecting the environment used for further processing. The training phase must reflect this changing environment at run time as well as provide classifications for each training instance based on the target output. These issues are discussed in the next section.

## 3.4 Creating the Training Instances

ID3 is a supervised learning algorithm that requires a set of training instances, each labeled with the correct classification for that instance. To create these instances Wrap-Up begins its tree building phase by passing the training texts to the sentence analyzer, which creates a set of objects representing the extracted information. Multiple references to the same object are then merged to form the initial input to Wrap-Up's first stage. Wrap-Up encodes instances and builds trees for this stage, then repeats the process using trees from stage one to build trees for stage two, and so forth until trees have been built for all six stages.

As it encodes instances, Wrap-Up repeatedly consults the target output to assign a classification for each training instance. When building trees for the Slot Filtering stage an instance is classified positive if the extracted information matches a slot in the target output. Consider the example of a reference to an "Ultratech stepper" in a microelectronics text. Sentence analysis creates an equipment object with two slots filled, equipment type stepper and equipment name "Ultratech". This stage of Wrap-Up has a separate ID3 tree to judge the validity of each slot, equipment type and equipment name.

Suppose that the target output has an equipment object with type "stepper" but that "Ultratech" is actually the manufacturer's name and not the equipment model name. The equipment type instance will be classified positive and the equipment name instance classified negative since no equipment object in the target output has the name Ultratech.

Does this instance include features that capture why a human analyst would not consider "Ultratech" to be the equipment name? The human is probably using world knowledge to recognize Ultratech as a familiar company name and recognize other names such as "Precision 5000" as familiar equipment names. Knowledge such as lists of known company names and known equipment names is not presently included in Wrap-Up, although this could be derived easily from the training corpus.

To create training instances for the second stage of Wrap-Up, the entire training corpus is processed again, this time discarding some slot values as spurious according to the Slot Filtering trees before creating instances for Slot Merging trees. An instance is created for each pair of objects of the same type. If both objects can be mapped to the same object in the target output, the instance is classified as positive. For example, an instance would be created for a pair of device objects, one with device type RAM and the other with size 256 KBits. It is a positive instance if the output has a single device object with type RAM and size 256 KBits.

By the time instances are created for later stages of Wrap-Up, errors will have crept in from previous stages. Errors in filtering, merging, and linking will have resulted in some objects retained that no longer match anything in the target output and some objects that only partially match the target output. Since some degree of error is unavoidable, it is best to let the training instances reflect the state of processing that will occur later when Wrap-Up is used to process new texts. If the training is too perfectly filtered, merged, and linked, it will not be representative of the underlying probabilities during run time use of Wrap-Up.

In later stages of Wrap-Up objects may become complex and only partially match anything in the target output. To aid in matching complex objects, one slot for each object type is identified in the output structure definition as the key slot. An object is considered to match an object in the output if the key slots match. Thus an object with a missing equipment name or spurious equipment name will still match if equipment type, the key slot, matches. If object A has a pointer to an object B, the object matching A in the output must also have a pointer to an object matching B.

Such recursive matching becomes important during the Link Creation stage. Among the last links considered in microelectronics are the roles a company plays towards a process. A company may be the developer of an x-ray lithography process that uses the ABC stepper, but not developer of the x-ray lithography process linked to a different equipment object. Wrap-Up needs to be sensitive to such distinctions in classifying training instances for trees in the Link Creation and Object Splitting stages.

Instances in the Inferring Missing Objects stage and the Inferring Missing Slot Values stage have classifications that go beyond a simple positive or negative. An instance for the Inferring Missing Objects stage is created whenever an object is found during training that has no higher object pointing to it. If a matching object indeed exists in the target output, Wrap-Up classifies the instance with the type of the object that points to it in the output. For example a training text may have a reference to "stepper" equipment, but have no mention of any process that uses the stepper. The target output will have a lithography object of type "unknown" that points to the stepper equipment. This is a legitimate inference to make, since steppers are a type of lithography equipment. The instance for the orphaned stepper equipment object will be classified as "lithography-unknown-equipment". This classification gives Wrap-Up enough information during run time to create the appropriate object.

An instance for Inferring Missing Slot Values is created whenever a slot is missing from an object which has a closed class of possible values, such as the "status" slot for equipment objects, that has the value of "in-use" or "in-development". When a matching object is found in the target output, the actual slot value is used as the classification. If the slot is empty or no such object exists in the output, the instance is classified as negative. As in the Inferring Missing Objects stage, negative is the most likely classification for many trees.

Next we consider the effects of tree pruning and confidence thresholds that can make the ID3 more cautious or more aggressive in its classifications.

## 3.5 Confidence Thresholds and Tree Pruning

With any machine learning technique there is a tendency toward "overfitting", making generalizations based on accidental properties of the training data. In ID3 this is more likely to happen near the leaf nodes of the decision tree, where the partition size may grow too small for ID3 to select features with much predictive power. A feature chosen to discriminate among half a dozen training instances is likely to be particular to those instances and not useful in classifying new instances.

The implementation of ID3 used by Wrap-Up deals with this problem by setting a pruning level and a confidence threshold for each tree empirically. A new instance is classified by traversing the decision tree from the root node until a node is reached where the partition size is below the pruning level. The classification halts at that node and a classification of positive is returned if the proportion of positive instances is greater than or equal to the confidence threshold.

A high confidence threshold will make an ID3 tree cautious in its classifications, while a low confidence threshold will allow more positive classifications. The effect of changing the confidence threshold is more pronounced as the pruning level increases. With a large enough pruning level, nearly all branches will terminate in internal nodes with confidence somewhere between 0.0 and 1.0. A low confidence threshold will classify most of these instances as positive, while a high confidence threshold will classify them as negative.

Wrap-Up automatically sets a pruning level and confidence threshold for each tree using tenfold cross-validation. The training instances are divided into ten sets and each set is tested on a tree built from the remaining nine tenths of the training. This is done at various settings to find settings that optimize performance.

The metrics used in this domain are "recall" and "precision", rather than accuracy. Recall is the percentage of positive instances that are correctly classified, while precision is the percentage of positive classifications that are correct. A metric which combines recall and precision is the f-measure, defined by the formula $f = (\beta^2 + 1)PR/(\beta^2 P + R)$ where $\beta$ can be set to 1 to favor balanced recall and precision. Increasing or decreasing $\beta$ for selected trees can fine-tune Wrap-Up, causing it to select pruning and confidence thresholds that favor recall or favor precision.

We have seen how Wrap-Up automatically derives the classifiers needed and the feature set for each classifier, and how it tunes the classifiers for recall/precision balance. Now we will look at the requirements for using Wrap-Up, with special attention to the issue of manual labor during system development.

## 3.6 Requirements of Wrap-Up

Wrap-Up is a domain-independent architecture that can be applied to any domain with a well defined output structure, where domain objects are represented as case frames and relationships between objects are represented as pointers between objects. It is appropriate for any information extraction task in which it is important to identify logical relationships between extracted information. The user must supply Wrap-Up with an output definition listing the domain objects to be extracted. Each output object has one or more slots, each of which may contain either extracted information or pointers to other objects in the output. One slot for each object is labeled as the key slot, used during training to match extracted objects with objects in the target output.

If the domain and application are already well defined, a user should be able to create such an output definition in less than an hour. For a new application, whose information needs are not established, there is likely to be a certain amount of trial and error in developing the desired representation. This need for a well defined domain is not unique to discourse processing or to trainable components such as Wrap-Up. All IE systems require clearly defined specifications of what types of objects are to be extracted and what relationships are to be reported.

The more time consuming requirement of Wrap-Up is associated with the acquisition of training texts and most importantly, hand-coded target output. While hand-coded targets represent a labor-intensive investment on the part of domain experts, no knowledge of natural language processing or of machine learning technologies is needed to generate these answer keys, so any domain expert can produce answer keys for use by Wrap-Up. A thousand microelectronics texts were used to provide training for Wrap-Up. The actual number of training instances from these training texts varied considerably for each decision tree. Trees that handled the more common domain objects had ample training instances from only two hundred training texts, while those that dealt with the less frequent objects or relationships were undertrained from a thousand texts.

It is easier to generate a few hundred answer keys than it is to write down explicit and comprehensive domain guidelines. Moreover, domain knowledge implicitly present in a set of answer keys may go beyond the conventional knowledge of a domain expert when there are reliable patterns of information that transcend a logical domain model. Once available, this corpus of training texts can be used repeatedly for knowledge acquisition at all levels of processing.

The architecture of Wrap-Up does not depend on a particular sentence analyzer or a particular information extraction task. It can be used with any sentence analyzer that uses keywords and local linguistic patterns for extraction. The output representation produced by Wrap-Up could either be used directly to generate database entries in a MUC-like task or could serve as an internal representation to support other information extraction tasks.

## 3.7 The Joint Ventures Domain

After Wrap-Up had been implemented and tested in the microelectronics domain, we tried it on another domain, the MUC-5 joint ventures domain. The information to be extracted in this domain are companies involved in joint business ventures, their products or services, ownership, capitalization, revenue, corporate officers, and facilities. Relationships between companies must be sorted out to identify partners, child companies, and subsidiaries. The output structure is more complex than that of microelectronics, with back-pointers, cycles in the output structure, redundant information, and longer chains of linked objects.

Figure 6 shows a text from the joint ventures domain and a diagram of the target output. With all the pointers and back-pointers, the output for even a moderately complicated text becomes difficult to understand at a glance. This text describes a joint venture between a Japanese company, Rinnai Corp., and an unnamed Indonesian company to build a factory in Jakarta. A tie-up is identified with Rinnai and the Indonesian company as partners and a third company, the joint venture itself, as a child company. The output includes an "entity-relationship" object which duplicates much of the information in the tie-up object. A corporate officer, the amount of capital, ownership percentages, the product "portable cookers", and a facility are also reported in the output.

RINNAI CORP., JAPAN'S LEADING GAS APPLIANCE MANUFACTURER, WILL SET UP A JOINT VENTURE IN INDONESIA IN AUGUST TO PRODUCE PORTABLE COOKERS FOR LOCAL USERS, PRESIDENT SUSUMU NAITO SAID MONDAY.

THE NEW FIRM WILL BE CAPITALIZED AT ONE MILLION DOLLARS, OF WHICH RINNAI IS SCHEDULED TO PUT UP 50 PCT AND A LOCAL DEALER 50 PCT, HE SAID.

IT WILL MANUFACTURE 3,000 TO 4,000 UNITS A MONTH INITIALLY AT A PLANT IN A 26,000-SQUARE-METER SITE IN JAKARTA, NAITO SAID, ADDING RINNAI AIMS TO START FULL-SCALE PRODUCTION NEXT SPRING.

THE NAGOYA-BASED COMPANY HAS NOW SEVEN OVERSEAS PRODUCTION UNITS.

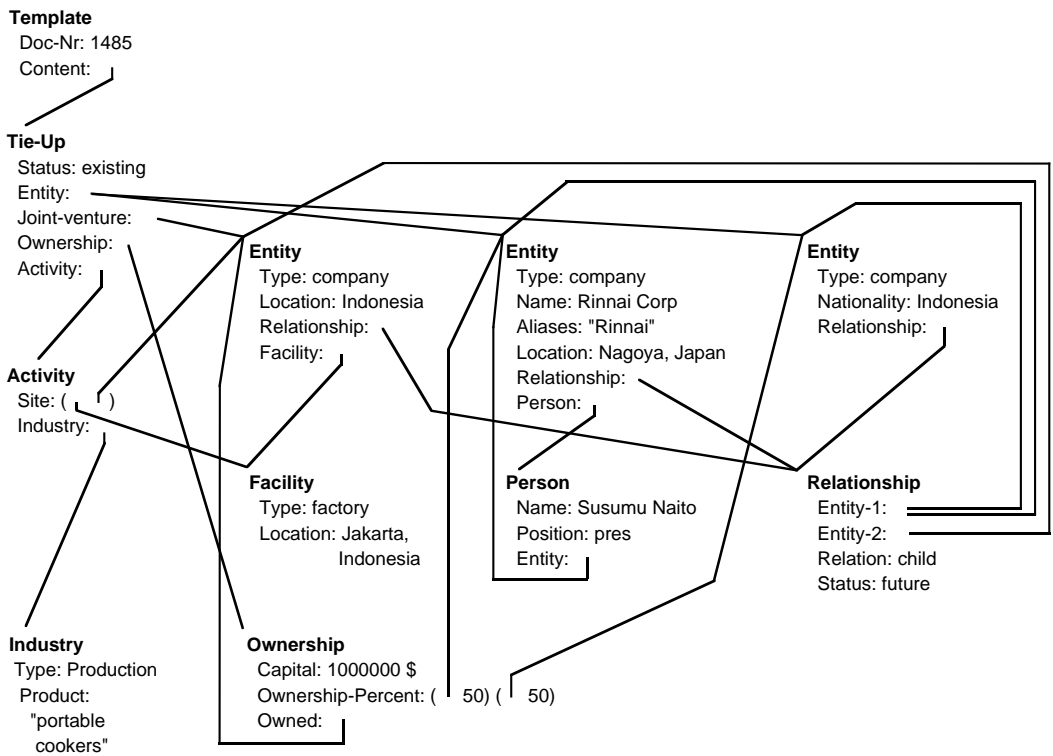

Figure 6: A sample text and target output from the joint ventures domain.

Some special handling was required for the joint ventures domain since the output structure defined for the MUC-5 evaluation included some slots such as activity site and ownership percent whose values had a mixture of extracted information and pointers. These slot values have their own internal structure and can be thought of as pseudoobjects, an activity site object with pointers to a facility object and a company, and an ownership percent object with a pointer to a company and another slot giving a numeric value. These pseudoobjects were reformulated as standard objects conforming to the requirements of Wrap-Up, the activity site slot pointing to an activity site object and so forth. These were then transformed back into the complex slot fills when printing the final representation of the output.

The output specifications for joint ventures were less well-behaved in other ways, with graph cycles, back pointers, and redundant objects whose content must agree with information elsewhere in the output. Modifications to Wrap-Up were needed to relax some implicit requirements for the domain structure, allowing graph cycles and giving special handling to any pointer slot which the user has labeled in the output definition as a back pointer.

Joint ventures also has some implicit constraints on relationships between objects. A company can play only a single role in a tie-up or a joint venture relationship: it cannot be both a joint venture child and also a parent or partner company. Wrap-Up had difficulty learning this constraint and performed better when certain pointer slots were labeled with a "single-role" constraint in the output definition.

This strategy of letting the user indicate constraints by annotating slots in the output definition was implemented in an ad hoc fashion. A more general approach would allow the user to declare several types of constraint on the output. A pointer slot may be required or optional, may have at most one pointer or allow several. Some slots of an object may be mutually exclusive, an entry in one prohibiting an entry in another slot. There may be a required agreement between the value of a slot in one object and a slot in another object. A fully domain-independent discourse tool needs a mechanism to implement such generalized constraints.

## 4. System Performance

As a point of comparison for the performance of Wrap-Up, the UMass/Hughes system was run with the TTG discourse module, which had been used in the official MUC-5 evaluation. Overall system performance with Wrap-Up was compared to performance with TTG, holding the rest of the system constant.

Wrap-Up takes the idea of TTG and extends it into a fully trainable system. TTG used decision trees to acquire domain knowledge, but often relied on hand-coded heuristics to apply that acquired knowledge, in particular the decisions about splitting or merging objects, which Wrap-Up handles during its Object Splitting stage; inferring missing objects, which Wrap-Up does in its Inferring Missing Objects stage; and adding context sensitive default slot values, which Wrap-Up does in its Inferring Missing Slot Values stage.

Several iterations of hand tuning were required to adjust thresholds for the decision trees produced by TTG, whereas Wrap-Up found thresholds and pruning levels to optimize recall and precision for each tree automatically. After a day of CPU time devoted to decision tree training, Wrap-Up produced a working system and no further programming was needed.

The comparison with TTG was made for both the microelectronics domain and the joint ventures domain. The metrics used here are recall and precision. Recall is the percentage of possible information that was reported. Correctly identifying two out of five possible company names gives recall of 40. Precision is the percent correct of the reported information. If four companies are reported, but only two of them correct, precision is 50. Recall and precision are combined into a single metric by the f-measure, defined as $f = (\beta^2 + 1)PR/(\beta^2 P + R)$, with $\beta$ is set to 1 for balanced recall and precision.

## 4.1 The Microelectronics Domain

Wrap-Up's scores on the official MUC-5 microelectronics test sets were generally a little higher than to those of TTG, both in overall recall and precision.

|        | Wrap-Up | | | TTG | | |
|--------|------|-------|------|------|-------|------|
|        | Rec. | Prec. | F    | Rec. | Prec. | F    |
| Part 1 | 32.3 | 44.4  | 37.4 | 27.1 | 39.5  | 32.1 |
| Part 2 | 36.3 | 38.6  | 37.4 | 32.7 | 37.0  | 34.7 |
| Part 3 | 34.6 | 37.7  | 36.1 | 34.7 | 40.5  | 37.5 |
| Avg.   | 34.4 | 40.2  | 36.8 | 31.5 | 39.0  | 34.8 |

Figure 7: Performance on MUC-5 microelectronics test sets

To put these scores in perspective, the highest scoring systems in the MUC-5 evaluation had f-measures in the high 40's. This was a difficult task both for sentence analysis and discourse analysis.

Another way to assess Wrap-Up is to measure its performance against the baseline provided by output from sentence analysis. Lack of coverage by the sentence analyzer places a ceiling on performance at the discourse level. In test set part 1 there were 208 company names to be extracted. The CIRCUS analyzer extracted a total of 404 company names, with only 131 correct and 2 partially correct, giving a baseline of 63% recall and 33% precision for that slot. Wrap-Up's Entity-Name-Filter tree managed to discard a little over half of the spurious company names, keeping 77% of the good companies. This resulted in 49% recall and 44% precision for this slot, raising the f-measure by 5 points, but doing so at the expense of recall.

Limited recall for extracted objects is compounded when it comes to links between objects. If half the possible companies and a third of the microelectronics processes are missing, discourse processing has no chance at a large proportion of the possible links between companies and processes.

Although precision is often increased at the expense of recall, Wrap-Up also has mechanisms to increase recall slightly. When the Inferring Missing Objects stage infers a missing process from an equipment object or the Object Splitting stage splits a process that points to multiple equipment, Wrap-Up can sometimes gain recall above that produced by the sentence analyzer.

## 4.2 The Joint Ventures Domain

In the joint ventures domain Wrap-Up's scores on the MUC-5 test sets were a little lower than the official UMass/Hughes scores. Wrap-Up tended to have lower recall but slightly higher precision.

|        | Wrap-Up | | | TTG | | |
|--------|------|-------|------|------|-------|------|
|        | Rec. | Prec. | F    | Rec. | Prec. | F    |
| Part 1 | 23.5 | 52.9  | 32.5 | 26.0 | 53.9  | 35.1 |
| Part 2 | 22.7 | 53.6  | 31.9 | 26.0 | 52.1  | 34.7 |
| Part 3 | 23.3 | 51.4  | 32.1 | 27.7 | 49.7  | 35.6 |
| Avg.   | 23.2 | 52.7  | 32.2 | 26.5 | 52.0  | 35.1 |

Figure 8: Performance on MUC-5 joint ventures test sets

The performance of Wrap-Up and TTG is roughly comparable for each of the two domains. Both systems tend to favor the domain in which they were first developed, Wrap-Up developed in microelectronics then ported to joint ventures, while the opposite was true for TTG. A certain amount of bias has probably crept into design decisions that were meant to be domain independent in each system. The higher scores of TTG for joint ventures are partly due to hand-coded heuristics that altered output from TTG before printing the final output, something that was not done for TTG in microelectronics or for Wrap-Up in either domain.

The most noticeable difference between Wrap-Up and TTG output in the joint ventures domain was in the filtering of spuriously extracted company names. Discourse processing started with 38% recall and 32% precision from sentence analysis for company names. Both systems included a filtering stage that attempted to raise precision by discarding spurious companies, but did so at the expense of discarding some valid companies as well. Each system used threshold settings to control how cautiously or aggressively this discarding is done (as in the example from Section 3.5). TTG's were set by hand and Wrap-Up's were selected automatically by cross-validation on the training set. TTG did only mild filtering on this slot, resulting in a gain of 2 precision points but a drop of 6 recall points. Wrap-Up chose aggressive settings and gained 13 precision points but lost 17 points in recall for this slot.

As a result, Wrap-Up ended up with only two thirds as many correct companies as TTG. This in turn meant two thirds as many pointers to companies in tie-ups and entity relationships. For other objects Wrap-Up scored higher recall than TTG, getting more than three times the total recall for activity, industry, and facility objects.

## 5. Conclusions

With the recent accessibility of large on-line text databases and news services, the need for information extraction systems is growing. Such systems go beyond information retrieval and create a structured summary of selected information contained within relevant documents. This gives the user the ability to skim vast amounts of text, pulling out information on a particular topic. IE systems are knowledge-based, however, and must be individually tailored to the information needs of each application.

Some research laboratories have focused on sophisticated user interfaces to ease the burden of knowledge acquisition. GE's NLToolset is an example of this approach (Jacobs et al., 1993), while BBN typifies systems that combine user input with corpus-based statistics (Ayuso et al., 1993). The University of Massachusetts has been moving in the direction of machine learning to create a fully trainable IE system. The ultimate goal is a turnkey system that can be tailored to new information needs by users who have no special linguistic or technical expertise.

Wrap-Up embodies this goal. The user defines an information need and output structure, and provides a training corpus of representative texts with hand-coded target output for each text. Wrap-Up takes it from there and instantiates a fully functional IE discourse system for the new domain with no further customization needed by the user. Wrap-Up is the first fully trainable system to handle discourse processing, and it does so with no degradation in performance. It automatically decides what classifiers are needed based on the domain output structure and derives the feature set for each classifier from sentence analyzer output.

The most intriguing aspect of Wrap-Up is the automatic generation of features. How effective was this, and what did the trees actually learn? The greatest leverage seems to come from features that encode attributes of domain objects. The trees in microelectronics often based their classification on probabilities conditioned on the device type, equipment type, or process type. The example tree in Section 3.2 first tested the equipment type and lithography type in determining whether a piece of equipment was used for a lithography process. This type of real world domain knowledge was the most important thing that Wrap-Up learned about microelectronics.

Useful knowledge was also provided by features that encoded the relative position of references in the text. Distance, measured in number of sentences apart, played a prominent role in many classifications, with other trees relying on more fine-grained features such as the number of times both references were in the same noun phrase or had overlapping linguistic context.

An enhancement to Wrap-Up's feature generation would be to increase its expressiveness about relative position. In addition to direct references to object A and object B, Wrap-Up could look for indirect references to A (pronominal or anaphoric) found near references to B and vice versa. The instance shown in Section 3.3 is an example where features for such indirect relationships might be useful.

Wrap-Up currently encodes an instance for each pair of objects that might be related, but is incapable of expressing the rule "attach object B to the most recent object of type A." It is blind to the existence of other objects that are alternate candidates to the relationship being considered. Features could be encoded to reflect whether object A is the most recently mentioned object of its type.

The features that were least successful and most tantalizing were those that encoded the local linguistic context, the extraction patterns. These included an exact lexical item and were nearly all of such low frequency that they added noise more often than aiding useful discriminations. Tree pruning was only a partial solution, and an experiment in combining semantically similar terms only caused a sharp drop in classification accuracy.

Low frequency terms are a built-in problem for any system that processes unrestricted text. Dunning (93) estimated that 20-30% of typical English news wire reports are composed of words of frequency less than one in 50,000 words. Yet the discourse decisions made by a human reader often seem to hinge on the use of one of these infrequent terms. It is a challenging open question to find methods to utilize local linguistic context without drowning in the noise produced by low frequency terms.

Finding a mechanism for choosing appropriate features is more critical than which machine learning algorithm is applied. ID3 was chosen as easy to implement, although other approaches such as vector spaces are worth trying. It is not obvious, however, how to craft a weighting scheme that gives greatest weight to the most useful features in the vector space and nearly zero to those not useful in making the desired discrimination. Cost and Salzberg (1993) describe a weighting scheme for the nearest neighbor algorithm that looks promising for lexically-based features. Another candidate for an effective classifier is a back propagation network, which might naturally converge on weights that give most influence to the most useful features.

We hope that Wrap-Up will inspire the machine learning community to consider analysis of unrestricted text as a fruitful application for ML research, while challenging the natural language processing community to consider ML techniques for complex processing tasks. In a broader context, Wrap-Up provides a paradigm for user customizable system design, where no technological background on the part of the user is assumed. A fully functional system can be brought up in a new domain without the need for months of development time, signifying substantial progress toward fully scalable and portable natural language processing systems.

## Appendix A: Walk-through of a Sample Text

To see the Wrap-Up algorithm in action, consider the sample text in Figure 9. The desired output has the company, Mitsubishi Electronics America, Inc., linked as purchaser/user to two packaging processes, TSOP and SOJ packaging. Each of these processes point to the device, 1 Mbit DRAM. The packaging material, plastic, should be attached to TSOP but not SOJ. All other details from the text are considered extraneous to the domain.

After sentence analysis, followed by the step that merges multiple references, there are eight objects passed as input to Wrap-Up. Sentence analysis did fairly well in identifying the relevant information, only missing "1 M" as a reference to 1 MBits. Three of the eight objects are spurious and should be discarded during Wrap-Up's Slot Filtering stage. According to domain guidelines, the name "Mitsubishi Electronics America, Inc." should be reported, not "The Semiconductor Division ...". The packaging material EPOXY and the device MEMORY should also be discarded.

The Slot Filtering stage creates an instance for each slot of each object. The Entity-Name-Filter tree classifies "Mitsubishi Electronics America, Inc." as a positive instance, but "The Semiconductor Division ..." as negative and it is discarded. The most reliable discriminator of valid company names is "extraction-count", which was selected as root feature of this tree. Training instances participating in several extraction patterns were twice as likely to be valid as those extracted only once or twice. This held true in this text.

The Semiconductor Division of Mitsubishi Electonics America, Inc. now offers
1M CMOS DRAMs in Thin Small-Outline Packaging (TSOP*), providing the
highest memory density available in the industry.  Developed by Mitsubishi,
the TSOP also lets designers increase system memory density with standard and
reverse or "mirror image," pin-outs.  Mitsubishi's 1M DRAM TSOP provides
the density of chip-on-board but with much higher reliability because the
plastic epoxy-resin package allows each device to be 100% burned-in and fully
tested.  *Previously referred to as VSOP (very small-outline package) or USOP
(ultra small-outline package).  The 1M DRAM TSOP has a height of 1.2 mm, a
plane measurement of 16.0 mm x 6.0 mm, and a lead pitch of 0.5 mm, making
it nearly three times thinner and four times smaller in volume than the 1M
DRAM SOJ package.  The SOJ has a height of 3.45 mm, a plane dimension of
17.15 mm x 8.45 mm, and a lead pitch of 1.27 mm.  Additionally, the TSOP
weighs only 0.22 grams, in contrast with the 0.75 gram weight of the SOJ.

Full text available on PTS New Product Announcements.

Figure 9: A microelectronics text

**Entity**
  Type: company
  Name:Mitsubishi Electronics
      America, Inc.

**Entity**
  Type: company
  Name: The Semiconductor Division of
      Mitsubishi Electronics America, Inc.

**Device**
  Type: DRAM

**Packaging**
  Type: TSOP

**Device**
  Type: MEMORY

**Packaging**
  Material: EPOXY

**Packaging**
  Material: PLASTIC

**Packaging**
  Type: SOJ

Figure 10: Input to Wrap-Up from the sample text

"Mitsubishi Electronics America, Inc." had extraction count of 5, while the spurious name was extracted from only 2 patterns.

As the Slot Filtering stage continues, the packaging material EPOXY is classified negative by the Packaging-Material-Filter tree, whose root test is packaging type. It turns out that EPOXY was usually extracted erroneously in the training corpus. This contrasts with the material PLASTIC which was usually reliable and is classified positive. Both TSOP and SOJ packaging types are classified positive by the Packaging-Type-Filter tree. Instances for these types were usually positive in the training set, particularly when extracted multiple times from the text. The Device-Type-Filter tree, with root feature device type, finds that DRAM is a reliable device type but that MEMORY was usually spurious in the training corpus. It should usually be merged with a more specific device type.

The Slot Merging stage of Wrap-Up then considers each pair of remaining objects of the same type. There are three packaging objects, one with type TSOP, one with material PLASTIC, and one with type SOJ. The Packaging-Slotmerge tree easily rejects the TSOP-SOJ instance, since packaging objects never had multiple types in training. After testing that the second object has no packaging type, the feature "distance" is tested. This led to a positive classification for TSOP-PLASTIC, which are from the same sentence, and negative for SOJ-PLASTIC, with nearest references two sentences apart. At this point four objects remain:

| Entity | Packaging |
|---|---|
| Type: company | Type: TSOP |
| Name:Mitsubishi Electronics America, Inc. | Material: PLASTIC |

| Device | Packaging |
|---|---|
| Type: DRAM | Type: SOJ |

The Link Creation stage considers each pair of objects that could be linked according to the output structure. The first links considered are pointers from packaging to device objects. Separate instances for the Packaging-Device-Link tree are created for the possible TSOP-DRAM link and for the possible SOJ-DRAM link. Although only 25% of the training instances were positive, the tree found that 78% were positive with packaging type TSOP and "distance" of 0 sentences, and 77% were positive with packaging type SOJ and device type DRAM. After testing a few more features, the tree found each of these instances positive and pointers were added in the output. Notice how this tree interleaves knowledge about types of packaging and types of devices with knowledge about relative position of references in the text.

The next Link Creation decision concern the roles Mitsubishi plays towards each of the packaging processes. The output structure has a "microelectronics-capability" object with one slot pointing to a lithography, layering, etching, or packaging process, and four other slots (labeled developer, manufacturer, distributor, and purchaser/user) pointing to companies. Wrap-Up accordingly encodes four instances for Mitsubishi and TSOP packaging, one for each possible role. The same is done for Mitsubishi and SOJ packaging.

Instances for Mitsubishi in the roles of developer, manufacturer, and distributor were all classified as negative. Training instances for these trees had almost no positive instances.

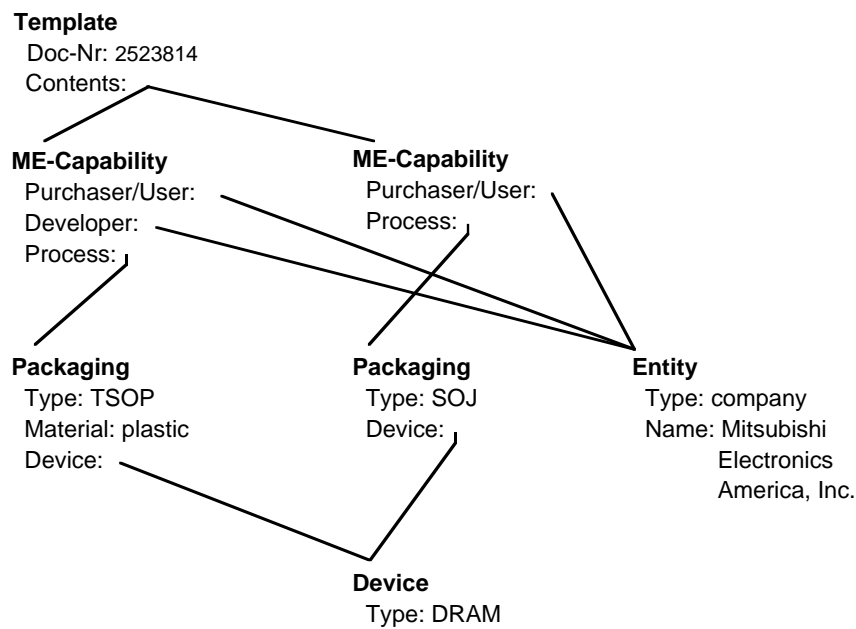

Figure 11: Final output after links have been added

It seems that stories about packaging processes in this corpus are almost exclusively about companies purchasing or using someone else's packaging technology.

There are seldom explicit linguistic clues about the relationship of a company to a process in this corpus, so the Packaging-User-Link tree tests first for the relative distance between references. Only 20% of training instances were positive, but when distance was 0 it jumped to 43% positive. Mitsubishi is in the same sentence with TSOP and the Mitsubishi-SOJ instance also has distance of 0 by inheritance. Even though the nearest reference to SOJ is two sentences after Mitsubishi, SOJ is linked to DRAM which occurs in the same sentence as Mitsubishi. Both instances are classified positive after further testing for packaging type and other features.

The last discourse decision in the Link Creation stage is to add pointers to each microelectronics capability from a "template object", created as a dummy root object in this domain's output. The Object Splitting stage finally gets to make a decision, albeit a vacuous one, and decides to let the template object point to multiple objects in its "content" slot. There were no "orphan" objects or missing slot values for the last two stages of Wrap-Up to consider. The final output for this text is shown in Figure 11.

## Acknowledgements

This research was supported by NSF Grant no. EEC-9209623, State/Industry/University Cooperative Research on Intelligent Information Retrieval.